\documentclass[letterpaper]{article} 
\usepackage[preprint]{aaai2027}  
\usepackage[hyphens]{url}  
\usepackage{graphicx} 
\usepackage{natbib}  
\usepackage{caption} 
\usepackage{amsmath}
\usepackage{amssymb}
\usepackage{booktabs}
\newcommand{\method}{ADMIL}

\newcommand{\includeorplaceholder}[3]{%
  \IfFileExists{#1}{%
    \includegraphics[width=\linewidth]{#1}%
  }{%
    \fbox{%
      \parbox[c][#2][c]{0.94\linewidth}{%
        \centering
        \textbf{Figure placeholder}\\[4pt]
        #3\\[4pt]
        \texttt{\detokenize{#1}}%
      }%
    }%
  }%
}

\newcommand{\includeorplaceholderlimited}[4]{%
  \IfFileExists{#1}{%
    \includegraphics[width=\linewidth,height=#2,keepaspectratio]{#1}%
  }{%
    \fbox{%
      \parbox[c][#3][c]{0.94\linewidth}{%
        \centering
        \textbf{Figure placeholder}\\[4pt]
        #4\\[4pt]
        \texttt{\detokenize{#1}}%
      }%
    }%
  }%
}

\title{ADMIL: Attention-Distilled Multiple Instance Learning \\for Selective Foundation Model Inference in Pathology}

\author{
    Duncan Stothers\textsuperscript{\rm 1}, Ren-Chin Wu\textsuperscript{\rm 2}, William Lotter\textsuperscript{\rm 1,3,4}\footnote{Correspondence: lotterb@ds.dfci.harvard.edu}
}
\affiliations{
    \textsuperscript{\rm 1} Department of Data Science, Dana-Farber Cancer Institute, Boston, MA \\
    \textsuperscript{\rm 2} Department of Pathology, Dana-Farber Cancer Institute, Boston, MA \\
    \textsuperscript{\rm 3} Department of Pathology, Brigham and Women’s Hospital, Boston, MA \\
    \textsuperscript{\rm 4} Harvard Medical School, Boston, MA
}

\begin{document}

\maketitle

\begin{abstract}
Attention-based multiple instance learning (ABMIL) using pathology foundation model embeddings is effective for slide-level tasks, but exhaustive inference requires applying a large image encoder to every foreground tile despite the subsequent attention distribution often concentrating over a small subset of informative regions. We introduce \textbf{ADMIL} (\textbf{A}ttention-\textbf{D}istilled \textbf{M}ultiple \textbf{I}nstance \textbf{L}earning), a selective-compute framework that distills an ABMIL teacher's attention into a lightweight tile-selection model, PriorNet. Using an EfficientNet architecture, PriorNet learns the teacher attention distribution from raw tile pixels with KL divergence; at inference, it scores the foreground pool, selects the top-$K$ tiles, and invokes the expensive foundation model only on that subset before a selected-bag ABMIL student predicts the slide label. Across BRACS, PANDA, and CAMELYON16, ADMIL matches full-teacher headline performance at $K=4$, $8$, and $128$ tiles, respectively, avoiding >98\% of foundation model (Virchow2) tile embeddings and model inference FLOPs. Random and teacher-attention oracle controls show that this result depends on task-relevant selection rather than tile-count reduction alone. Quantitative and qualitative analyses suggest that PriorNet recovers the teacher's tile ordering with high fidelity while focusing on task-relevant morphological regions.  
ADMIL shows that nearly all expensive tile encodings can be removed without sacrificing slide-level performance, providing a potential path for more efficient deployment in clinical settings where latency and compute costs are key considerations. 
\end{abstract}

\begin{figure*}[t]
  \centering
  \includegraphics[width=0.95\textwidth]{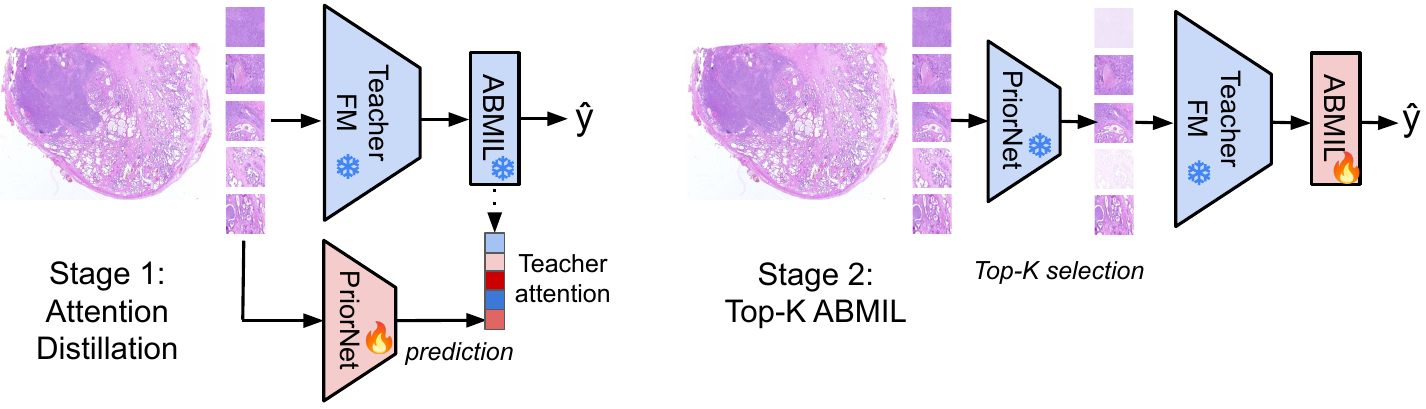}
  \caption{
    \textbf{ADMIL overview.}
    Stage 1: A lightweight image encoder (PriorNet) learns to estimate the attention distribution produced by a frozen foundation model-based ABMIL teacher. Stage 2: PriorNet selects the top-$K$ tiles for generating foundation model embeddings, which are used to train a separate top-$K$ ABMIL predictor. At deployment, ADMIL reduces compute cost by decreasing the number of foundation model calls. 
  }
  \label{fig:method}
\end{figure*}

\section{Introduction}

Pathology foundation models have become a standard source of tile-level representations for whole-slide image (WSI) analysis. A typical pipeline partitions each WSI into foreground tiles, embeds every tile with a foundation model encoder, and aggregates the resulting bag with attention-based multiple-instance learning (ABMIL) \citep{ilse2018attention, campanella2019clinical, lu2021clam, chen2024uni, lu2024conch, vorontsov2024virchow, xu2024gigapath}. This paradigm has shown strong performance for a range of tasks, including metastasis detection, tumor subtyping, and histologic grading \citep{campanella2019clinical, litjens2018camelyon, brancati2022bracs, bulten2022panda, diab2025}. Its computational profile, however, is poorly matched to clinical deployment. A single WSI may contain thousands to tens of thousands of foreground tiles, so exhaustive foundation model inference can dominate latency, compute costs, and throughput.

This cost is particularly striking because trained ABMIL models rarely use all tiles equally. Their slide prediction is often supported by a highly concentrated attention distribution over a small subset of morphologically informative regions for the particular task. The expensive encoder is therefore applied broadly even though the final classifier concentrates on a narrow evidence set. This mismatch suggests a selective-compute strategy: use a cheap model to identify likely high-value tiles before invoking the expensive pathology foundation model.

We introduce Attention-Distilled Multiple Instance Learning (\method{}), a framework to substantially reduce the inference cost of foundation model-based ABMIL while preserving performance. Given a trained ABMIL model, ADMIL first distills its attention distribution into a lightweight encoder, PriorNet, which is trained using KL divergence to predict the teacher-assigned attention weight for each tile. For a user-defined deployment budget $K$, a separate ABMIL is then trained using foundation model embeddings from only the $K$ tiles receiving the highest PriorNet scores. At inference, PriorNet scores the full foreground pool, the foundation model embeds only the selected tiles, and the corresponding top-$K$ ABMIL model produces the slide prediction.


This design differs from conventional model compression. PriorNet does not replace the foundation model or the ABMIL architecture; it gates access to the expensive encoder. It also differs from heuristic tissue filtering and random subsampling because the selection rule is distilled from task-specific teacher behavior. The resulting pipeline preserves the high-capacity representation where it matters while avoiding most foundation model evaluations elsewhere.

Our contributions are threefold:
\begin{itemize}
    \item We formulate attention distillation as a selective-compute policy for foundation model-based ABMIL and show that a lightweight PriorNet trained by KL divergence recovers useful teacher evidence.
    \item Across BRACS, PANDA, and CAMELYON16, ADMIL matches the full-teacher headline performance at small tile budgets, avoiding $98.5\%$--$99.8\%$ of foundation model tile embeddings and reducing model inference FLOPs by $98.3\%$--$99.6\%$.
    \item We quantitatively and qualitatively evaluate tile selection fidelity, observing strong concordance between teacher- and student-selected tiles. 
\end{itemize}

\section{Related Work}

\paragraph{Weakly supervised WSI classification.}
WSI classification is naturally formulated as multiple-instance learning, where a slide is a bag of tiles and supervision is available only at the bag level. Attention-based MIL introduced a permutation-invariant learned aggregation rule consisting of an interpretable, attention-weighted average of tile features \citep{ilse2018attention}. Large-scale studies and later variants such as CLAM, DSMIL, and TransMIL demonstrated the effectiveness of weakly supervised WSI learning \citep{campanella2019clinical, lu2021clam, li2021dsmil, shao2021transmil}. When combined with foundation models, ABMIL remains a standard approach and often still outperforms other variants \citep{shao2025do}. ADMIL is nonetheless complementary to potential improvements in the aggregator: it targets the upstream cost of producing foundation model features for every candidate tile.

\paragraph{Pathology foundation models.}
Self-supervised and multimodal pathology models have become strong general-purpose feature extractors. CTransPath and HIPT introduced transformer-based and hierarchical pretraining for histopathology \citep{wang2022ctranspath, chen2022hipt}; UNI and CONCH demonstrated broad transfer across visual and multimodal tasks respectively \citep{chen2024uni, lu2024conch}; and Virchow and Prov-GigaPath further scaled pathology pretraining and whole-slide modeling \citep{vorontsov2024virchow, xu2024gigapath}. These models improve downstream performance but make exhaustive tile encoding increasingly expensive. ADMIL is designed specifically for this high-capacity, high-cost feature-extraction regime.

\paragraph{Efficiency for gigapixel pathology.}
Several prior works have sought to reduce the computational burden of WSI analysis through new MIL strategies. ZoomMIL introduced a differentiable coarse-to-fine MIL network that progressively identifies informative regions and allocates higher-resolution processing only to selected areas \cite{zoommil}. HDMIL similarly distinguishes between high- and low-resolution processing by distilling high-resolution attention into a low-resolution pre-screening network \cite{dong2025hdmil}. Recent work has specifically targeted the cost of pathology foundation models. H0-mini focuses on distilling the features of large foundation models directly, resulting in a more efficient encoder \cite{h0mini}. EAGLE uses a cheaper WSI-level foundation model to select a fixed set of 25 tiles for processing by a larger foundation model \cite{eagle}. Rather than pursuing efficiency through multi-resolution processing (e.g., ZoomMIL, HDMIL), compressed feature extraction (e.g., H0-mini), or task-agnostic tile selection using a separately pretrained model (e.g., EAGLE), ADMIL distills a task-specific foundation model ABMIL teacher directly into a lightweight attention predictor. The resulting PriorNet ranks tiles before foundation model inference, retaining the original high-capacity encoder for informative regions while avoiding its application to the vast majority of the slide.


\section{Methods: ADMIL}

\begin{figure*}[t]
  \centering
  \includegraphics[width=\textwidth]{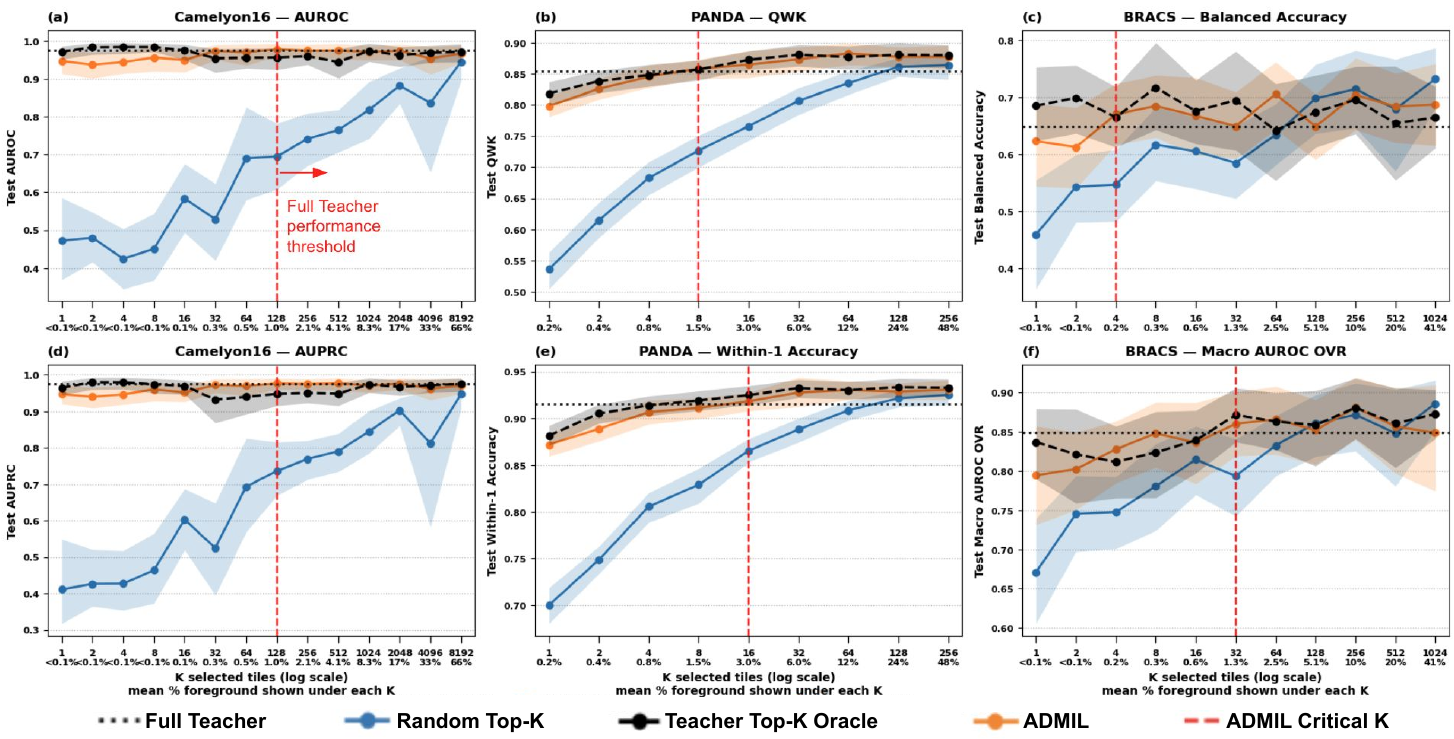}
  \caption{\textbf{Slide-level performance versus selected tile budget.} Columns correspond to CAMELYON16, PANDA, and BRACS. The top row reports each dataset's primary metric and the bottom row its secondary metric. The black dotted line denotes full-teacher performance, which is compared to ADMIL, a random tile selector, and a top-$K$ teacher oracle. The red dashed vertical line indicates the $K$ value where ADMIL reaches teacher performance. The x-axis is logarithmic and reports both raw $K$ and the corresponding mean foreground fraction. Error bars correspond to 95\% confidence intervals.} 
  \label{fig:headline-curves}
\end{figure*}

\begin{figure*}[t]
  \centering
  \includegraphics[width=\textwidth]{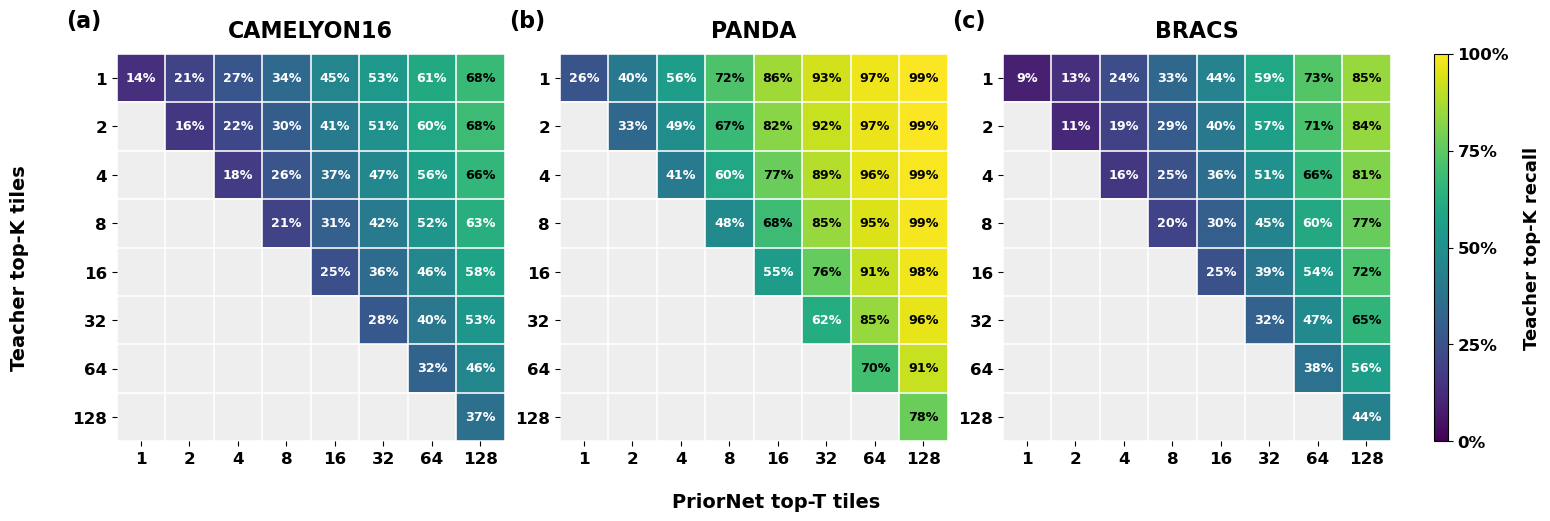}
  \caption{\textbf{Teacher-support retrieval by PriorNet.} Values correspond to the fraction of the teacher top-$K$ tiles found among the PriorNet top-$T$ tiles, averaged across test slides within each dataset. Cells with $T<K$ are masked. All panels use the same color scale.}
  \label{fig:retrieval}
\end{figure*}

\begin{figure*}[h]
  \centering
  \includegraphics[width=\textwidth]{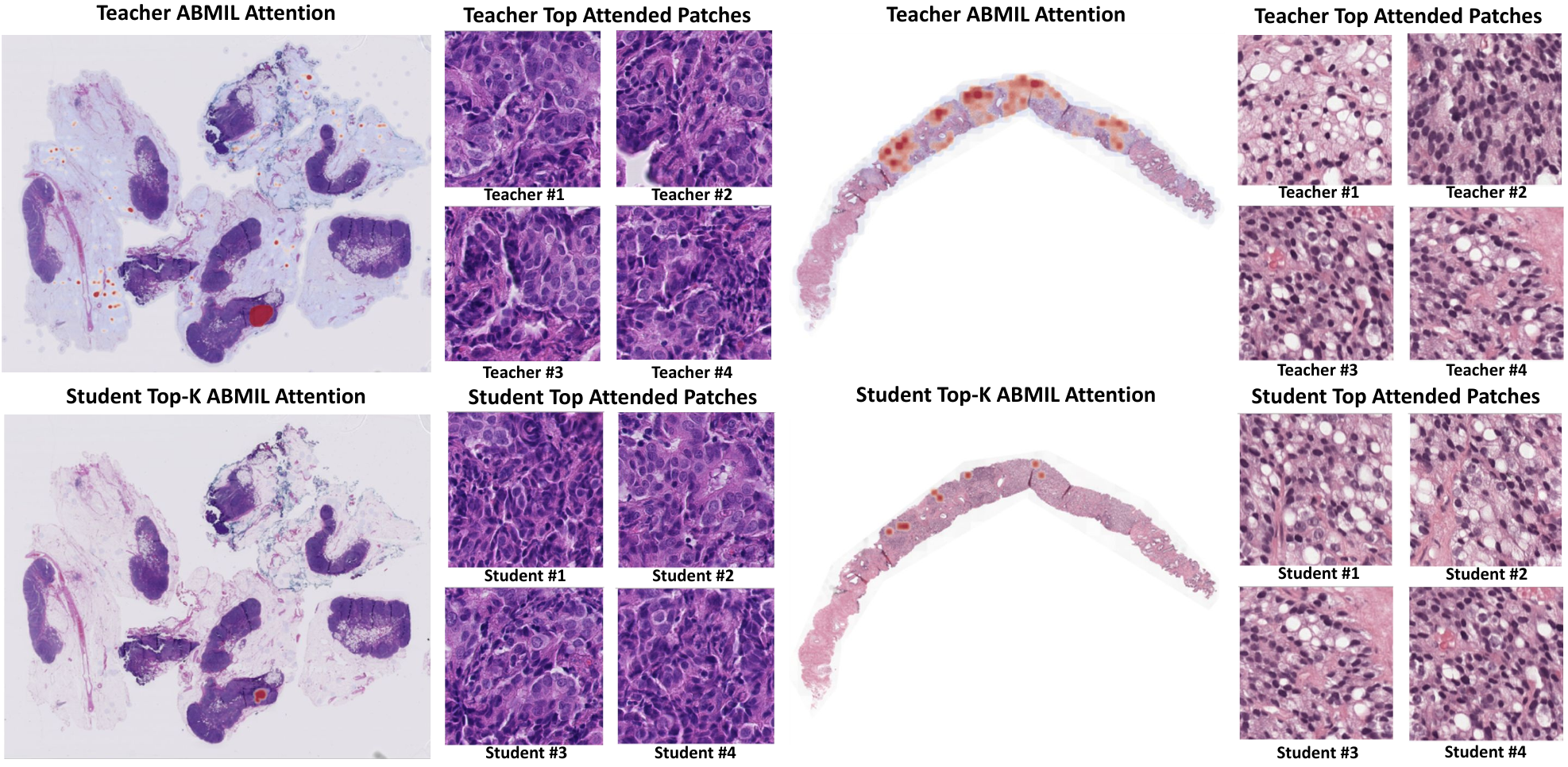}
  \caption{\textbf{Example attention maps.} Each row contains a representative test slide from one dataset (left: CAMELYON16, right: PANDA). Panels show the WSI thumbnail, full-bag teacher attention, selected-bag student attention, and the top attended tiles. }
  \label{fig:qualitative}
\end{figure*}

\begin{figure*}[t]
  \centering
  \includegraphics[width=\textwidth]{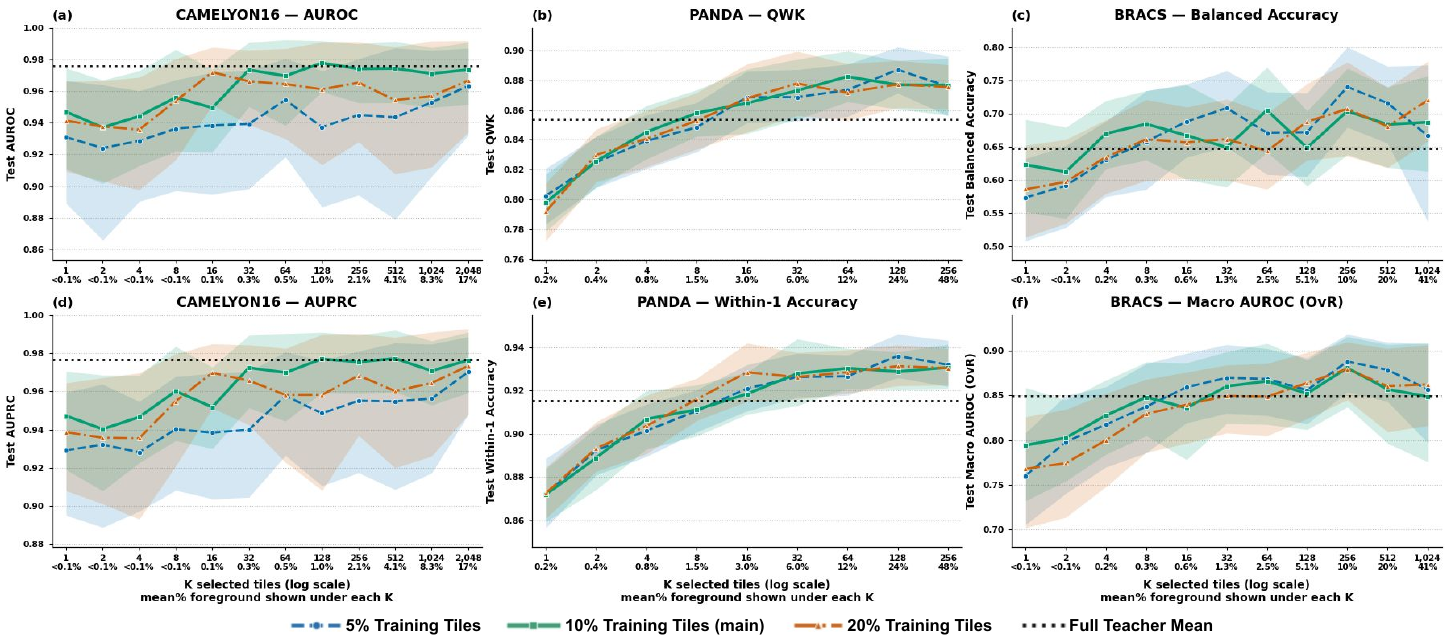}
  \caption{\textbf{Sensitivity to the PriorNet supervision tile budget.} Each panel compares PriorNet models trained with 5\%, 10\%, or 20\% of each slide's foreground candidates. The black dotted line denotes the full-teacher reference. Error bars correspond to 95\% confidence intervals. The 10\% setting is the default used for main results; the other curves assess robustness to the supervision fraction.}
  \label{fig:sensitivity}
\end{figure*}

\subsection{Problem Setting and Full-Bag Teacher}

A slide is represented by a foreground tile bag $X = \{x_i\}_{i=1}^{N}$, where $N$ varies by slide. A frozen foundation model $f_\theta$ maps each tile to an embedding $z_i=f_\theta(x_i)$. The full-bag teacher computes attention logits
\[
a_i = w_a^\top \tanh(W_a z_i+b_a)
\]
and normalized attention
\[
\alpha_i = \frac{\exp(a_i)}{\sum_{j=1}^{N}\exp(a_j)}.
\]
The teacher slide representation is $h=\sum_i \alpha_i z_i$, followed by a task head $g_\psi(h)$. The teacher is trained on the full foreground bag and then frozen. Its exported attention distribution provides tile-level supervision for PriorNet.

\subsection{Attention Distillation with KL Divergence}

PriorNet is a lightweight image-space model, $p_\phi$,
that produces one scalar score $s_i=p_\phi(x_i)$ per foreground tile. Scores are normalized across the slide:
\[
q_i = \frac{\exp(s_i)}{\sum_{j=1}^{N}\exp(s_j)}.
\]
PriorNet is trained with the KL distillation objective,
\[
\mathcal{L}_{\mathrm{prior}}
=
D_{\mathrm{KL}}(\boldsymbol{\alpha}\,\|\,\mathbf{q})
=
\sum_{i=1}^{N}
\alpha_i\log\frac{\alpha_i+\epsilon}{q_i+\epsilon}.
\]

Rather than training PriorNet uniformly over all tiles, we use attention-guided sampling to increase the representation of highly attended tiles while reducing training cost. Specifically, for each dataset, we train on $Z\%$ of foreground tiles in the training set, distributed evenly across WSIs to target an equal number of tiles, $M$, per WSI. The retained tiles per WSI consist of the top $M/2$ ranked by teacher attention and an additional $M/2$ sampled from the remaining foreground pool. For the KL distillation loss, the teacher's attention distribution is re-normalized over the sampled tiles. 
In the main experiments, we set $Z=10\%$ and additionally assess sensitivity using $Z \in \{5,20\}\%$. During evaluation, deployment, and top-$K$ ABMIL training, PriorNet scores the complete foreground tile pool.

\subsection{Top-$K$ Selection and Selected-Bag ABMIL}

Let $\pi_s$ denote the permutation that orders PriorNet scores from largest to smallest. ADMIL selects
\[
S_K(X)=\{\pi_s(1),\ldots,\pi_s(\min(K,N))\}.
\]
Only selected tiles are embedded by the foundation model at inference:
\[
z_i=f_\theta(x_i),\qquad i\in S_K(X).
\]
A separate top-$K$ ABMIL is trained under this selected-bag condition. For $i\in S_K(X)$, it computes
\[
\rho_i=w_b^\top\tanh(W_b z_i+b_b),
\qquad
\beta_i=\frac{\exp(\rho_i)}{\sum_{j\in S_K(X)}\exp(\rho_j)},
\]
then forms
\[
h_K=\sum_{i\in S_K(X)}\beta_i z_i,
\qquad
\hat y_K=g_\omega(h_K),
\]
with optimization using the original slide-level task loss. 


\section{Methods: Evaluation Approach}
We compare ADMIL against several baselines, evaluating predictive performance, inference efficiency, and agreement between PriorNet predictions and teacher ABMIL attention values. 

\subsection{Baselines and Controls}

\paragraph{Full foundation model ABMIL teacher.}
The original ABMIL teacher embeds every foreground tile using the foundation model and serves as the full-compute performance reference.

\paragraph{Random top-$K$.}
For each slide, $K$ foreground tiles are sampled uniformly without replacement and used to train and evaluate a separate selected-bag ABMIL architecture. This controls for whether any subset of the same size is sufficient.

\paragraph{Teacher top-$K$ oracle.}
The oracle uses the teacher's own full-bag attention to select
\[
S_K^{\mathrm{oracle}}(X)=\operatorname{TopK}(\{\alpha_i\}_{i=1}^{N}).
\]
It is not deployable because obtaining $\alpha$ requires full-bag foundation model inference, but it estimates the performance attainable when teacher evidence is recovered perfectly.

\subsection{Attention Fidelity Analysis}

PriorNet's ability to recapitulate the teacher's attention is evaluated using correlation and retrieval metrics. For every slide, we compute the Pearson and Spearman correlation between PriorNet and teacher normalized attention values. We then average these values across test-set slides to obtain dataset-level agreement metrics. We also measure teacher-support retrieval: for teacher support size $K$ and PriorNet retrieval depth $T\geq K$, we compute
\[
R_\ell(K,T)
=
\frac{
\left|
\operatorname{TopK}(\boldsymbol{\alpha}_\ell)
\cap
\operatorname{TopT}(\mathbf q_\ell)
\right|
}{\min(K,N_\ell)}.
\]
We evaluate $K,T\in\{1,2,4,8,16,32,64,128\}$ and mask cells where $T<K$. This analysis compares full-foreground teacher attention $\alpha$ with the full-foreground PriorNet distribution $q$; it is distinct from the selected-bag student attention $\beta$.

\subsection{Sensitivity Analysis}

We assess sensitivity to the training tile sampling parameter $Z$. The default $Z = 10\%$ setting is used for all headline results; the $5\%$ and $20\%$ settings test whether the method's behavior depends strongly on this supervision fraction. All other training and evaluation procedures are held fixed across settings. This analysis is intended to assess robustness to $Z$, rather than to select an alternative headline configuration.

\subsection{Inference Efficiency Assessment}
Inference efficiency is assessed based on percentage of foundation model embeddings avoided and FLOPs saved. These calculations are reported for a critical budget $K_{\mathrm{crit}}$, which corresponds to the number of tile embeddings needed to reach teacher performance for a particular dataset and task. Details of these calculations are provided in the Appendix.  

\section{Methods: Experimental Setup}

\subsection{Datasets and Tasks}

We evaluated ADMIL on three public WSI benchmarks. CAMELYON16 contains lymph node biopsies for breast cancer metastasis detection \citep{litjens2018camelyon, bejnordi2017camelyon}, with 270 WSIs for model development and 129 for testing and an average of 12,391 foreground tiles per WSI. PANDA consists of prostate biopsies for six-class
ISUP grade classification \citep{bulten2022panda} with an average of 531 foreground tiles per WSI. As PANDA does not have a canonical test split, we split the data into 90/10\% development/testing corresponding to 9,553 and 1,062 WSIs, respectively. BRACS contains breast biopsies and we assess the three-class benign, atypical, and malignant
classification task \citep{brancati2022bracs}, with 458 and 87 WSIs for development and testing, respectively, with an average of 2,513 foreground tiles per slide.

We performed three global runs for each dataset using seeds 0, 1,
and 2. For each global seed, we kept the test partition fixed and
split the development set into a label-stratified 70/30
training/validation split using the global seed. PriorNet and all ABMIL training used the training split; model checkpoints were selected based on the validation split. PANDA was split at
the patient level because it contains multiple slides per patient; CAMELYON16 and BRACS only contain one slide per patient.

\subsection{Statistical Analysis}
We report AUROC and AUPRC for CAMELYON16, quadratic weighted kappa
(QWK) and within-1 accuracy for PANDA, and balanced accuracy and macro
one-vs-rest AUROC for BRACS. Unless otherwise noted, presented metrics correspond to average test-set performance across the three independent runs. 
Uncertainty is estimated with 1,000 hierarchical bootstrap replicates.

\subsection{Models and Training}

\paragraph{Model Architectures.}
We use Virchow2 as the foundation model tile encoder \citep{vorontsov2024virchow}. Full-bag teachers and selected-bag students use non-gated ABMIL with a 512-dimensional tanh attention layer, a scalar attention scorer, and a task head with a 512-dimensional hidden layer. PriorNet uses an EfficientNet-B0 \cite{efficientnet} backbone initialized with ImageNet-1K V1 weights and a scalar output head. All models operated on foreground-only tiles of size $224\times224$.

\paragraph{PriorNet Training.}
A separate PriorNet model was trained on each dataset for 20 epochs using AdamW with an initial learning rate of $10^{-4}$, batch size corresponding to 1 WSI, weight decay of $10^{-5}$, cosine learning-rate decay, mixed precision, dropout of 0.2, and gradient-norm clipping at 1.0. As described above, PriorNet was trained using a KL-divergence loss, with checkpoint selection based on the validation split. PriorNet training was independent of deployment budget $K$.

\paragraph{ABMIL Training.}
Teacher and student ABMIL models were optimized with AdamW for 40 epochs with weight decay of $10^{-4}$, batch size of 1, cosine learning rate decay, mixed-precision training, dropout of 0.1, and gradient-norm clipping at 1.0. Teachers used an initial learning rate of $10^{-4}$; students used an initial rate of $10^{-3}$, as we found that a higher learning rate facilitated convergence when training with fewer tiles. CAMELYON16 used class-balanced binary cross entropy as the loss function; PANDA and BRACS used class-balanced cross entropy. A separate ABMIL student was trained for every evaluated tile budget $K$. Checkpoints for all ABMIL models were selected based on the best validation headline metric during training.

\section{Results}

\subsection{ADMIL Matches Full-Bag Performance at Small Tile Budgets}

Figure~\ref{fig:headline-curves} shows the performance of ADMIL across tile budgets compared to the full-teacher ABMIL model. For each dataset, a low-budget regime is observed in which ABMIL matches teacher performance with a small percentage of foundation model embedded tiles. On the primary metrics, ADMIL first reaches the teacher threshold at $K=4$ for BRACS balanced accuracy, $K=8$ for PANDA QWK, and $K=128$ for CAMELYON16 AUROC. These budgets correspond to $99.8\%$, $98.5\%$, and $99.0\%$ fewer foundation model tile embeddings, respectively (Table~\ref{tab:headline}), which translates to similar levels of model inference FLOPs saved given PriorNet's lightweight architecture. The secondary metrics show the same qualitative behavior. ADMIL reaches the mean teacher AUPRC on CAMELYON16 at $K=128$ (0.977; 95\% CI [0.959, 0.991]), the mean teacher within-1 accuracy on PANDA at $K=16$ (0.918; 95\% CI [0.909, 0.928]), and the mean teacher macro one-vs-rest AUROC on BRACS at $K=32$ (0.860; 95\% CI [0.819, 0.899]). ADMIL's performance via PriorNet-based tile selection also consistently approaches the teacher-selection oracle and substantially exceeds random sampling, especially at low $K$.

\begin{table*}[t]
\centering
\caption{
Performance at the descriptive critical tile budget
$K_{\mathrm{crit}}$. FM saved indicates the percentage of foundation model tile embeddings avoided. Model FLOPs saved accounts for PriorNet, foundation model, and ABMIL operations at inference. 
}
\label{tab:headline}
\resizebox{\textwidth}{!}{%
\begin{tabular}{@{}llccccccc@{}}
\toprule
Dataset
& Metric
& $K_{\mathrm{crit}}$
& \shortstack{Full\\teacher}
& \shortstack{Teacher\\95\% CI}
& \shortstack{ADMIL at\\$K_{\mathrm{crit}}$}
& \shortstack{ADMIL\\95\% CI}
& \shortstack{FM saved (\%)}
& \shortstack{FLOPs saved (\%)} \\
\midrule

BRACS
& Balanced Acc.
& 4
& 0.648
& [0.594, 0.700]
& \textbf{0.670}
& [0.617, 0.720]
& 99.8
& 99.6 \\

PANDA
& QWK
& 8
& 0.854
& [0.838, 0.870]
& \textbf{0.858}
& [0.842, 0.874]
& 98.5
& 98.3 \\

CAMELYON16
& AUROC
& 128
& 0.976
& [0.954, 0.991]
& \textbf{0.978}
& [0.960, 0.992]
& 99.0
& 98.7 \\

\bottomrule
\end{tabular}%
}
\end{table*}

\subsection{PriorNet Recovers the Teacher's Attention Ordering}

\begin{table}[t]
\centering
\caption{Correlation between teacher attention and PriorNet scores. Values are computed per WSI and then averaged across WSIs.}
\label{tab:attention-correlation}
\resizebox{\linewidth}{!}{%
\begin{tabular}{@{}lcccc@{}}
\toprule
Dataset & Pearson & Pearson 95\% CI & Spearman & Spearman 95\% CI  \\
\midrule
CAMELYON16 & 0.456 & [0.425, 0.487] & 0.799 & [0.778, 0.821]  \\
PANDA & 0.622 & [0.532, 0.713] & 0.839 & [0.816, 0.861]  \\
BRACS & 0.342 & [0.246, 0.412] & 0.579 & [0.518, 0.628]  \\
\bottomrule
\end{tabular}%
}
\end{table}

The performance results suggest that PriorNet effectively identifies task-informative tiles by distilling the teacher model's attention. To directly assess how well PriorNet recapitulates the teacher's attention patterns, we computed the correlation between PriorNet-predicted and teacher-assigned tile attention using the test set of each dataset. PriorNet exhibits positive agreement with the teacher across all three datasets (Table~\ref{tab:attention-correlation}). Spearman correlation is consistently higher than Pearson correlation, suggesting that PriorNet more faithfully recovers the teacher's ranking of tile importance than the exact magnitude of its attention weights. Rank agreement is strongest on PANDA (0.839), followed by CAMELYON16 (0.799) and BRACS (0.579).

The retrieval matrices in Figure~\ref{fig:retrieval} provide a direct selection-oriented view of fidelity. Despite an average of 12,391, 2,513, and 531 foreground tiles per WSI for CAMELYON16, BRACS, and PANDA respectively, PriorNet's top selected tile matches the top teacher tile 14\%, 9\%, and 26\% of the time for the three datasets respectively, with recall increasing as the PriorNet retrieval depth grows. Together with the correlation results, this indicates that the lightweight model learns a useful approximation to the teacher's task-specific evidence ranking before foundation model embedding.

\subsection{Qualitative Review Finds Histologically Plausible Evidence}

Representative ABMIL-derived attention maps produced by the teacher and student models are shown in Figure~\ref{fig:qualitative}. In the CAMELYON16 example (left), both models assign high attention to tiles containing tumor cells, consistent with the ground-truth metastasis label and the corresponding model predictions. In the PANDA example (right), several of the highest-attention tiles overlap between the two models and contain high-grade prostate carcinoma morphology consistent with the ISUP 5 label. In both examples, the student attention map largely aligns with the teacher while showing higher sparsity, driven by the top-$K$ selection mechanism. Overall, this qualitative analysis provides a sanity check that the student attends to plausible morphology, complementing the quantitative analyses. 

\subsection{Performance Is Robust Across PriorNet Supervision Budgets}

Results of the sensitivity analysis for the training tile-sampling percentage are contained in Figure~\ref{fig:sensitivity}. Across datasets and metrics, the three curves follow similar performance trajectories with overlapping confidence intervals, suggesting robustness to this hyperparameter. We retain $10\%$ as the default and interpret the $5\%$ and $20\%$ runs as evidence that the method is not narrowly tuned to one exact fraction.

\section{Discussion}

Our results show that the standard foundation model-based ABMIL pipeline contains substantial redundant tile-level computation. The full teacher processes every foreground tile, yet ADMIL reaches the same mean headline performance using only four selected BRACS tiles, eight PANDA tiles, and 128 CAMELYON16 tiles. The corresponding reduction in foundation model calls exceeds $98\%$ in all three datasets. This is not simply a consequence of bag redundancy: random top-$K$ students perform worse in the low-budget regime, whereas the teacher-attention oracle and learned PriorNet selection retain much more of the full-bag performance.

The magnitude of the compute reduction follows from the large per-tile asymmetry between the selector and the foundation model. The Virchow2 checkpoint requires $340.132$ GFLOPs per tile, compared with $0.769$ GFLOPs for PriorNet. Scanning the complete foreground pool with PriorNet contributes only about $0.226\%$ of the full-teacher model-forward FLOPs, after which Virchow2 is applied to just $0.16\%$, $1.51\%$, and $1.03\%$ of foreground tiles on BRACS, PANDA, and CAMELYON16, respectively. Consequently, estimated model inference savings remain close to the reduction in Virchow2 calls despite scoring every foreground tile with PriorNet.

The attention-fidelity results clarify what PriorNet recovers. Spearman correlation is consistently higher than Pearson correlation, suggesting that exact probability calibration is less important than preserving the ordering of teacher-relevant tiles. Some variation across datasets is also observed, where higher correlations are observed for PANDA and CAMELYON16 than BRACS. BRACS combines heterogeneous benign, atypical, and malignant morphologies and uses a larger physical field of view, which may make local teacher attention harder to predict from isolated raw tiles. Nevertheless, BRACS still reaches the full-teacher balanced accuracy at only four selected tiles, suggesting that perfect global correlation is not necessary for effective decision-focused retrieval.

Sensitivity experiments show that ADMIL is not tied to one exact PriorNet supervision budget. The 5\%, 10\%, and 20\% curves retain the same qualitative shape and largely overlapping uncertainty, with small task-specific differences rather than a consistent ordering.

Compared to strong prior works focusing on hierarchical spatial processing \cite{zoommil, dong2025hdmil} or task-agnostic model compression or tile selection \cite{h0mini, eagle}, ADMIL is designed specifically for efficient inference with high-performing, MIL-level foundation model pipelines. The benefit of foundation model representations lies in their adaptability across downstream tasks; however, in clinical use, predictions are ultimately made for specific tasks, and the most relevant tiles can differ accordingly. ADMIL learns to approximate this relevance based on teacher supervision, enabling highly streamlined inference while preserving the underlying model representations and prediction architecture.

Several limitations remain. First, ADMIL depends on the stability and relevance of teacher attention. A high-performing teacher may attend to confounders or distribute evidence diffusely, reducing selector reliability. Second, $K_{\mathrm{crit}}$ is a post-hoc summary of the test curve rather than a deployable validation-selected operating point; a production system should choose $K$ on validation data. Third, we assessed ADMIL using three common benchmarks and a high-performing base foundation model, but future validation with additional datasets and FMs is an important direction. Finally, although ADMIL is designed to reduce inference time and computational cost for scalable deployment, it introduces additional overhead during training.

\section{Conclusion}

ADMIL converts a trained ABMIL teacher's attention into a selective-compute policy. With a KL-trained PriorNet and a selected-bag ABMIL student, the method preserves teacher performance while avoiding most foundation model embeddings across three distinct tasks. Quantitative attention agreement, teacher-support retrieval, supervision-budget sensitivity, and morphology-consistent examples jointly support the central mechanism: the teacher's implicit localization behavior can be distilled early enough to reduce expensive inference. This offers a simple and extensible path toward more deployable pathology foundation model systems.

\bibliography{aaai2027}

\section{Appendix}

\subsection{Inference Efficiency Assessment Details} 

For slide $\ell$ with $N_\ell$ foreground tiles, the fraction of expensive foundation model tile embeddings avoided at budget $K$ is
\[
\mathrm{FM\text{-}savings}(K)
=
1-
\frac{\sum_\ell \min(K,N_\ell)}{\sum_\ell N_\ell}.
\]
Calculation of FLOPs ($F$) for model inference additionally includes PriorNet scoring over the full foreground pool and the teacher/student MIL heads:
\[
F_{\mathrm{ADMIL},\ell}
=
N_\ell F_{\mathrm{PriorNet}}
+
\min(K,N_\ell)F_{\mathrm{FM}}
+
F_{\mathrm{topK MIL}}\]
compared with
\[
F_{\mathrm{teacher},\ell}
=
N_\ell F_{\mathrm{FM}}
+
F_{\mathrm{teacher\ MIL}}\]
I/O and preprocessing are excluded from FLOPs calculations, and thus reduction in FLOPs should be interpreted at the model-forward level rather than wall-clock latency or total computation.

For each dataset and metric, we define the descriptive critical budget
\[
K_{\mathrm{crit}}
=
\min\{K:\overline{M}_{\mathrm{ADMIL}}(K)\geq \overline{M}_{\mathrm{teacher}}\},
\]
where the bars denote means over the three observed global runs. $K_{\mathrm{crit}}$ is computed from the complete test performance curve and is therefore a post-hoc descriptive summary, not a validation-selected deployment hyperparameter. Checkpoint selection itself uses validation data only.

\end{document}